\documentclass[10pt, a4paper]{article}

\usepackage[]{lrec-coling2024} 

\usepackage{times}
\usepackage{latexsym}
\usepackage{graphicx}
\usepackage{booktabs}
\usepackage{multirow}
\usepackage{tabularx}
\usepackage{xspace}
\usepackage{array}
\usepackage{svg}
\usepackage{balance}

\newcommand\ours{\textsc{MACP}\xspace}

\title{ProCQA: A Large-scale Community-based Programming Question Answering Dataset for Code Search}

\name{Zehan Li$^{1,2}$, Jianfei Zhang$^{3}$, Chuantao Yin$^{2}$, Yuanxin Ouyang$^{3}$, Wenge Rong$^{1,3}$} 

\address{$^{1}$State Key Laboratory of Complex \& Critical
Software Environment, Beihang University, China \\
$^{2}$Sino-French Engineer School, Beihang University, China \\
$^{3}$School of Computer Science and Engineering, Beihang University, China \\
         \{lizehan, zhangjf, chuantao.yin, oyyx, w.rong\}@buaa.edu.cn}

\abstract{
Retrieval-based code question answering seeks to match user queries in natural language to relevant code snippets. Previous approaches typically rely on pretraining models using crafted bi-modal and uni-modal datasets to align text and code representations. In this paper, we introduce ProCQA, a large-scale programming question answering dataset extracted from the StackOverflow community, offering naturally structured mixed-modal QA pairs. To validate its effectiveness, we propose a modality-agnostic contrastive pre-training approach to improve the alignment of text and code representations of current code language models. Compared to previous models that primarily employ bimodal and unimodal pairs extracted from CodeSearchNet for pre-training, our model exhibits significant performance improvements across a wide range of code retrieval benchmarks.
\\ \newline \Keywords{Code QA Dataset, Code Search, Contrastive Pretraining} }

\begin{document}

\maketitleabstract

\section{Introduction}

Code Question Answering (Code QA) represents a pivotal research area in software intelligence.
One popular task formulation is retrieval-based QA~\citep{deepcodesearch}, in which the primary objective is to effectively match user queries expressed in natural language to relevant code snippets from an existing corpus.
The prevailing approach for retrieval-based Code QA has been the utilization of dual-encoder-based representation models.
The core idea underlying this approach is to map natural language queries and code snippets into a shared representation space, where closely located vectors correspond to semantically similar meanings.

To learn a shared representation space for text and code, early research efforts adopted masked language modeling (MLM) objective on paired text-code dataset to align different modalities~\citep{pmlr-v119-kanade20a, feng-etal-2020-codebert}, similar to the monolingual and cross-lingual pre-training approaches~\citep{devlin-etal-2019-bert, xlm}.
Subsequent work discovered the potential of contrastive pre-training, and applied it to code representation learning by constructing large-scale paired datasets~\citep{jain-etal-2021-contrastive, li-etal-2022-coderetriever}.

\begin{figure}
    \centering
    \includegraphics[width=0.45\textwidth]{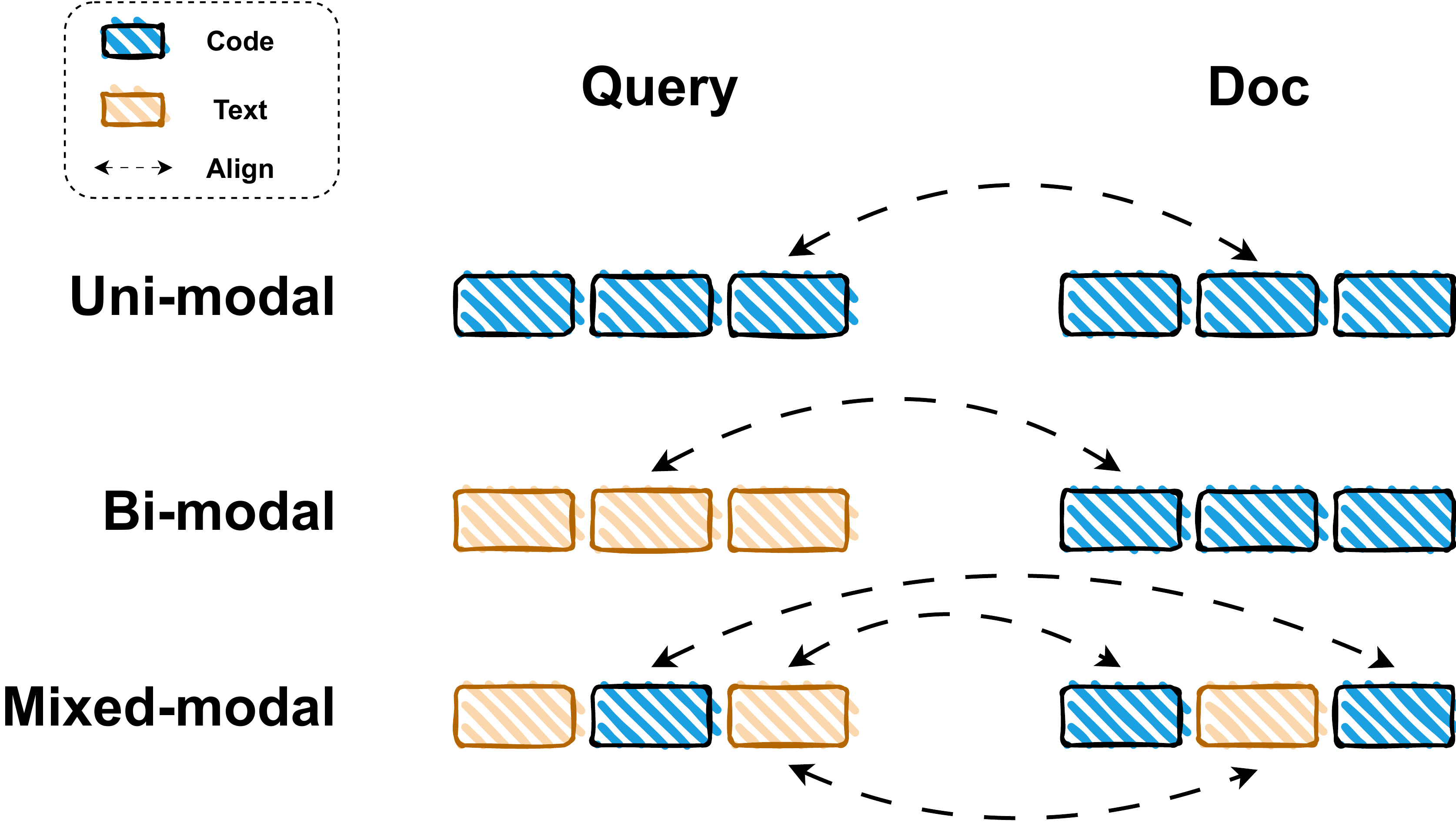}
    \caption{Illustration of different data formats used for contrastive representation alignment. Color represents chunk modality. Unimodal data focuses on code-to-code matching, while bimodal data emphasizes cross-modal matching. The mixed-modal data in ProCQA enables simultaneous learning of all matching patterns.}
    \label{fig:dataformat}
\end{figure}

Current contrastive code representation learning methods such as CodeRetriever~\citep{li-etal-2022-coderetriever} typically rely on curated uni-modal (code-code pairs) or bi-modal data (text-code pairs).
Few work even uses distinct encoder for text and code~\citep{heyman2020neural, code-pretraining}.
Such pre-training design emphasizes the concept of modality distinction, diverging from the goal of establishing a unified representation space for different modalities.
In Figure~\ref{fig:dataformat}, we illustrate different data formats used for contrastive pre-training and analyse their chunk-level matching patterns.
Uni-modal data offers code-code matching patterns, whereas bi-modal data implies code-text matching patterns.
While a combination of both data types during pre-training can enable models to learn both matching signals, a more data-efficient approach to capture all matching patterns is through mixed-modal data.

\begin{table*}[!ht]
    \centering
    \begin{tabularx}{\textwidth}{l|X}
    \toprule
    Question & Segmentation fault while copying a string to the memory allocated array \\
    \hline
    Description & Following is a program I am practicing;
int main() {

    int i = 0;
    char **grid = (char **) malloc(5*sizeof(int));
        for (i = 0 ; i < 5 ; i++)
            grid[i] = (char *) malloc(6);
            strcpy(grid[0], "eabcd");
            strcpy(grid[1], "fghij");
            strcpy(grid[2], "olkmn");
            strcpy(grid[3], "trpqs");
            strcpy(grid[4], "xywuv"); /*Segmentation Fault at this line*/
        return 0;
}
I am getting a segmentation fault at the line  strcpy(grid[4], "xywuv"); . What could be the reason? I have allocated the array to have 5 strings(rows) of 6 characters each(columns).
\\
    \hline
    Answer & You are allocating the wrong type at line 3
char **grid = (char **) malloc(5*sizeof(int));
Should be
char **grid = (char **) malloc(5*sizeof(char*));
This is because you are declaring string-array. Therefore, the malloc should be char* (string / character pointer)
Also the same if you were trying to declare 2-D integer array. It will be
int **grid = (int **) malloc(5*sizeof(int*));
 \\
    \bottomrule
    \end{tabularx}
    \caption{An example sampled from the C programming language subset of ProCQA. Text and code are interleaved in these QA pairs.}
    \label{tab:case-study}
\end{table*}

Besides, the majority of code embedding models~\citep{jain-etal-2021-contrastive, wang2021syncobert, li-etal-2022-coderetriever} have primarily relied on CodeSearchNet~\citep{husain2020codesearchnet} as the main pre-training corpus. While CodeSearchNet is a valuable resource, its size and data distribution have inherent limitations that may impact the quality and diversity of learned code representations.
Recent work has proposed to curate large-scale code datasets from GitHub~\citep{allal2023santacoder}.
Yet their efforts mainly focus on training large-scale generative language models (LMs).
In parallel, some research endeavors have aimed to create code-related question-answering datasets from diverse sources, as evidenced by~\citet{huang-etal-2021-cosqa, lee-etal-2022-cs1qa}.
Nevertheless, most of these datasets remain constrained by their scale, rendering them more suitable for stand-alone evaluation benchmarks rather than comprehensive pre-training corpus.

Therefore in this research, we try to bridge these gaps by proposing \textbf{ProCQA}, a large-scale community-based programming question answering dataset mined from StackOverflow.
ProCQA encompasses an extensive collection of approximately 5 million QA pairs, spanning 11 different programming languages.
This dataset is distinguished by its comprehensive language coverage, the diversity of user queries, and its code-mixing data format\footnote{Please refer to Table~\ref{tab:case-study} for an illustrative example.}. It can be used as both an evaluation benchmark and a pre-training corpus.
We provide strict rule-based filtering and data decontamination procedure to ensure its quality and fairness.
Different types of baseline models are trained and compared on this dataset to test its suitability as an evaluation benchmark.

To assess the efficacy of our proposed dataset as a pre-training corpus, we conducted large-scale modality-agnostic contrastive pretraining (\ours) on the code-mixing dataset, without making distinctions between text and code modalities.
To demonstrate whether \ours can learn a better aligned representation space, we evaluated it on extensive code retrieval benchmarks, covering supervised, zero-shot, and out-of-domain scenarios.
Experiments reveal that compared to previous pre-trained code language models, \ours achieves substantial improvements on most tasks we considered, advancing the previous best code retrieval model CodeRetriever~\citep{li-etal-2022-coderetriever} by 1$\sim$10\% points across different evaluation benchmarks.
Comprehensive ablation and analysis demonstrates the effectiveness of our proposed approach.

The contributions of this paper can be summarized as follows:
\begin{itemize}
    \item We create ProCQA, a large-scale dataset for programming question answering.
    ProCQA is characterized by its practicality, diversity and mixed-modal data format.
    We demonstrate its potential as an evaluation benchmark for comparing different code language models.
    \item Based on ProCQA, we present \ours, a code representation model pre-trained with modality-agnostic contrastive learning on the large-scale code-mixing dataset. \ours demonstrates remarkable performance gains over prior approaches across a wide range of code retrieval tasks.
\end{itemize}

\section{Related Work}

\subsection{Code QA}

Code-based question answering is a sub-problem of question answering.
Different from the generative formulation, retrieval-based code QA aims to retrieve the most similar code from a large-scale code corpus, satisfying user requests.
To evaluate the neural code search ability of current models, CodeSearchNet~\citep{husain2020codesearchnet} was constructed by mining large-scale comment-code pairs from public GitHub repositories.
Additionally, to evaluate the code comprehension ability of language models, \citet{liu-wan-2021-codeqa-question} introduced CodeQA, a free-form code question-answering dataset. This dataset was derived from existing code summarization datasets mined from GitHub, including two widely-used programming languages Python and Java. CodeQA synthesizes various types of question-answer pairs from code comments and documentation strings using manually curated rules, templates, and a range of NLP toolkits.

Recent work has been focused on constructing code QA dataset from real-world scenarios.
For example, CoSQA~\citep{huang-etal-2021-cosqa} mines real-world user queries from Bing search logs and utilizes models trained on CodeSearchNet and humans to label corresponding code.
Moreover, educational programming QA datasets have also gained attention.
CS1QA~\citep{lee-etal-2022-cs1qa} collects student questions and answers from teaching assistants on an online forum designed for an introductory Python programming course. This dataset offers insights into the educational applications of code-based question answering.

\subsection{Code Language Models}
Language models pre-trained on large-scale unlabeled corpora have demonstrated significant potential in code understanding and generation tasks.
Prior works such as CodeBERT~\citep{feng-etal-2020-codebert} employed replaced language modeling on uni-modal and bi-modal data for pre-training.
GraphCodeBERT~\citep{guo2021graphcodebert} advanced this approach by harnessing data flow encoded in the Abstract Syntax Tree (AST) of code to enrich code structural information during pre-training.
UniXCoder~\citep{guo-etal-2022-unixcoder} unified three pre-training designs into one architecture and utilized AST structure and code comment to enhance the cross-modal alignment.
There are also some work on adapting generative language models for code, as exemplified by CodeT5~\citep{wang-etal-2021-codet5} and PLBART~\citep{ahmad-etal-2021-unified}. These models incorporate code structure information into the design of specific pre-training tasks.
Contrastive methods have also been introduced into code pre-training by several recent works with different approaches proposed for constructing positive and negative pairs~\citep{jain-etal-2021-contrastive, wang2021syncobert, ding-etal-2022-towards, corder, li-etal-2022-coderetriever}.

It is worth noting that current code language models' pre-training corpus are primarily sourced from CodeSearchNet, consisting of 2 million code-text pairs. Limited efforts have been dedicated to mining large-scale datasets from GitHub~\citep{allal2023santacoder, li2023starcoder}, but they mainly focus on training decoder language models rather than code representation models.
An exception is the work by OpenAI~\citep{text-and-code}, but their models are only available via paid APIs and training data is not detailed.

\section{ProCQA}

In this section, we outline the methodologies employed in the creation of the ProCQA dataset, along with the filtering strategies applied to ensure data quality and fairness. Additionally, we present an analysis of various dataset statistics and define two tasks utilizing this dataset to evaluate different baseline models. The source code is available at \href{https://github.com/jordane95/procqa}{https://github.com/jordane95/procqa}.

\subsection{Data Acquisition}

To ensure the diversity and reflect real world user problems, we crawl our dataset from StackOverflow, a question answering community focusing on solving programming problems.
Users can post their problems on the website and wait for others' answers.
One characteristic of this dataset is that both the question and answer are code-mixing, i.e., text and code are interleaved within these fields.
Such data format is very useful to indoctrinate and evaluate the model's matching ability of different patterns.

We use the public dumps as of 2022/12 for raw data downloading\footnote{https://archive.org/details/stackexchange}.
We extract the textual content consisting of code and texts from XML files.
Three fields (title, question, answer) are kept.
HTML tags are removed and only text content are kept using BeautifulSoup library.

\subsection{Data Cleaning}

A critical problem with these QA communities is that there are many unanswered questions and wrong answers.
To handle this issue, we apply some rule-based approaches to filter out low-quality questions and answers.

More specifically, we filter out questions/answers that are either too short (< 20 characters) or too long (> 4096 characters).
We only keep questions that have answers marked as accepted by the questioner since it is a natural annotation signal indicating the answer is helpful for the user.

\subsection{Data Format}
Data in ProCQA is formatted as triples illustrated in Table~\ref{tab:case-study}.
The question is a concise user request.
It is coupled with a detailed description which explains the problem in more detail.
The answer is posted by other user and is the one accepted by the questioner.
Note that in all data fields, code and text are interleaved, which provides a natural supervision signal for aligning the two modalities.

\subsection{Data Statistics}

\begin{table*}[!ht]
    \centering
    \resizebox{\textwidth}{!}{
    \begin{tabular}{cccccccccccc}
        \toprule
        \textbf{PL} & C & C++ & Java & Python & Ruby & Lisp & JavaScript & C\# & Go & Rust & PHP \\
        \midrule
        \textbf{Size} & 204746 & 418346 & 831697 & 1008478 & 131218 & 4612 & 1217095 & 817970 & 36011 & 15514 & 567357 \\
        \bottomrule
    \end{tabular}
    }
    \caption{Number of QA pairs for each programming language in ProCQA.}
    \label{tab:pls}
\end{table*}

We partition the dataset into different programming language subsets according to their tags contained in meta information.
We consider the following eleven languages based on their popularity: Python, Java, JavaScript, Ruby, C, C++, C\#, Rust, PHP, Lisp and Go.
Dataset statistics are shown in Table~\ref{tab:pls}.
We split the dataset into train / valid / test set by a proportion of 80\%:10\%:10\% following chronological order of posting date.

\begin{figure}[!ht]
    \centering
    \includegraphics[width=0.46\textwidth]{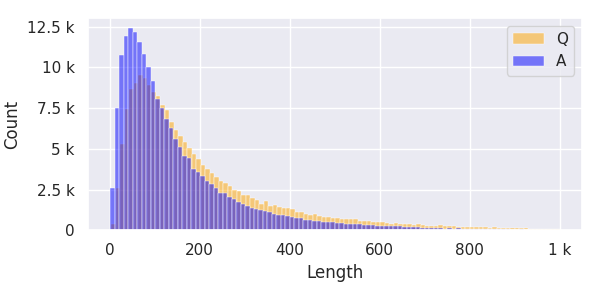}
    \caption{Question and answer length distribution in ProCQA (C subset).}
    \label{fig:qa_length}
\end{figure}

In addition, we analyse the question and answer length distribution of our ProCQA dataset in Figure~\ref{fig:qa_length}.
Most of the QA pairs in ProCQA contain dozens or hundreds of words, which are much closer to real user questions.

\subsection{Decontamination}
Since ProCQA is crawled from StackOverflow, it many overlap with some evaluation sets constructed from the same source.
To avoid data contamination, we perform evaluation data deduplication for our ProCQA training set.

Specifically, we employ two methods for deduplication.
The first one is based on substring matching.
Training example in ProCQA dataset is dropped if it contains any substring that is part of the queries in the evaluation set.
We use three evaluation sets to perform deduplication (CoNaLa, SO-DS and StaQC).
After this step, about 0.5\% examples from the Python subset are dropped.
Other subsets are influenced lightly.
We also apply fuzzy deduplication method based on MinHash but no additional duplicate is found.

\subsection{Comparison to previous datasets}
\begin{table*}[h]
\centering
\begin{tabular}{lclcc}
\toprule
\textbf{Dataset} & \textbf{\# of PLs}  & \textbf{Data Format}                   & \textbf{Size}    & \textbf{Data Source} \\ \midrule
CodeNN           & 2                   & {Title, code}                          & $\sim$187K pairs & StackOverflow        \\
CodeSearchNet    & 6                   & Comment, code                          & $\sim$2M pairs   & GitHub               \\
CodeQA           & 2                   & Question, answer, code                 & $\sim$190K pairs & GitHub               \\
CoSQA            & 1                   & Query, code                            & $\sim$20K pairs  & Web search           \\
CS1QA            & 1                   & Chat log, question, answer, type, code & $\sim$9K pairs   & Classroom            \\ \midrule
ProCQA           & 11                  & Question, description, answer          & $\sim$5M pairs   & StackOverflow        \\ \bottomrule
\end{tabular}
\caption{Comparison between different code-based datasets.}
\label{tab:code_datasets}
\end{table*}

To better understand the difference with previous dataset, we summarize some key factors of our ProCQA and previous ones in Table~\ref{tab:code_datasets}, including the number of supported programming languages (PLs), data format, size and data source.

CodeNN~\citep{iyer-etal-2016-summarizing} is also a dataset mined from StackOverflow for code summarization but contains much smaller amount of training examples and languages.
CodeSearchNet (CSN) is on pair with ProCQA in terms of languages and size but drawn from a different data distribution (GitHub). Its queries are either documentation strings or comments rather than natural language questions, limiting its practicality in real scenarios.
CoSQA and CS1QA contain some real user queries collected from Bing logs and classrooms but only cover Python and are limited in size.

In summary, ProCQA differs from previous work in the following main aspects:

1. More diverse language distribution at a larger scale.

2. Long-form questions and answers more aligned with real-world scenarios.

\subsection{Tasks}
We define two tasks based on the collected dataset for pilot exploration, including answer retrieval and generation.
We choose C subset as a test bed for comparing multiple language models.

\paragraph{Answer Retrieval}
This task is defined as finding the correct answer from a large-scale answer corpus.
We use answers from all splits of the dataset to form retrieval corpus.
The query is the concatenation of question and description.
We choose BM25 and some recent neural language models, such as BERT~\citep{devlin-etal-2019-bert}, CodeBERT~\citep{feng-etal-2020-codebert} and UniXCoder~\citep{guo-etal-2022-unixcoder}.
Neural LMs are fine-tuned with the contrastive learning objective (\emph{i.e.}, InfoNCE loss) on the question answer pairs from the training set.
All models are trained for 3 epochs with the batch size of 32 and the learning rate of 2e-5.
Both questions and answers are truncated to be maximum of 256 tokens.
We choose MRR@10, Recall@10 and Recall@100 as main evaluation metrics.

\begin{table}[!ht]
    \centering
    \begin{tabular}{lcccc}
    \toprule 
    Model & MRR@10 & R@10 & R@100 \\
    \midrule
    BM25 & 51.7 & 61.1 & 73.1 \\
    BERT & 48.3 & 62.0 & 79.7 \\
    CodeBERT & 53.0 & 66.8 & 83.5 \\
    UniXCoder & 58.4 & 71.8 & 86.1 \\
    \bottomrule
    \end{tabular}
    \caption{Answer retrieval performance of different language models on the C subset of ProCQA.}
    \label{tab:retrieval_qa_c}
\end{table}

Results are demonstrated in Table~\ref{tab:retrieval_qa_c}.
We observe that text-only language models such as BERT are even inferior to unsupervised BM25, in terms of MRR@10.
With code-specific pre-training, CodeBERT can outperform the strong BM25 baseline.
More recent code language models such as UniXCoder performs best on this task.

%

\paragraph{Answer Generation}
We also consider a generative task formulation, in which the model is required to directly generate the answer to the question without additional reference.
Similarly, we benchmark several generative language models on this task.
Selected baseline models include T5~\citep{t5}, CodeT5~\citep{wang-etal-2021-codet5}, PLBART~\citep{ahmad-etal-2021-unified}.
Models are trained in a sequence-to-sequence manner by optimizing the cross-entropy loss of the answer sequence given question sequence with the same training hyperparameters as stated above.
During inference, beam search decoding is used with a beam size of 5.
We use ROUGE~\cite{lin-2004-rouge} as main evaluation metrics for this task and demonstrate results in Table~\ref{tab:genqa_c}.

\begin{table}[!ht]
    \centering
    \begin{tabular}{lccc}
    \toprule 
    Model & ROUGE$_1$ & ROUGE$_2$ & ROUGE$_L$ \\
    \midrule
    T5 & 14.3 & 2.6 & 11.8 \\
    CodeT5 & 17.6 & 4.8 & 14.0 \\
    PLBART & 19.9 & 5.9 & 15.3 \\
    \bottomrule
    \end{tabular}
    \caption{Answer generation results of different baselines on ProCQA (C subset).}
    \label{tab:genqa_c}
\end{table}

It is found that code language models is better than text-only models, indicating the effectiveness of code-specific pre-training.
Even the best model struggles on this task because the answers are relatively long (mostly 100-200 words, see Figure~\ref{fig:qa_length}).
This indicates that ProCQA is a challenging dataset for long-form generative QA task.
How to improve the long-form question-answering performance of language models with limited parameters is also an interesting direction for future research.

\section{Experiments}
To assess the quality and utility of our proposed dataset, we evaluate its benefits to other code search benchmarks when acting as a pre-training corpus.
We also conduct ablation experiments to demonstrate the effectiveness of ProCQA over existing pre-training corpus CSN.

\subsection{Settings}

Our model basically follows the two-tower architecture in which vector representations for code and text are produced by mean pooling over the last layer hidden representations of the language models. It is trained via the contrastive objective using the InfoNCE loss
\begin{equation}
\label{eq:loss}
\mathcal{L} = - \log \frac{e^{s(q, d)/\tau}}{e^{s(q, d)/\tau} + \sum\limits_{d'\in\mathcal{D}_-}e^{s(q, d')/\tau}}.
\end{equation}
where $q$ denotes the question, $d$ denotes the corresponding answer, $\mathcal{D}_-$ is a set of negative samples, $\tau$ is the temperature.
$\mathcal{D}_-$ is also enlarged with other examples from the same batch.

The main baselines we compare to are GraphCodeBERT~\citep{guo2021graphcodebert} and CodeRetriever~\citep{li-etal-2022-coderetriever}. In addition to the text-code pairs in CSN used by GraphCodeBERT, CodeRetriever also employs sophisticated rules and learned models for mining high-quality code-code and code-text pairs from the raw CSN code corpus.
Instead we use commonly available QA pairs from ProCQA mined by weak supervision.
We use the training split across all languages to construct different types of mixed-modal positive pairs.
We apply modality-agnostic contrastive pre-training on the ProCQA and CSN dataset and compare our model to previous code embedding models on various retrieval tasks.
Our model is denoted as \ours.

\subsection{Implementation Details}
For fair comparison, our model is initialized with GraphCodeBERT, same as CodeRetriever.
\ours is pre-trained with the contrastive objective in Equation~\ref{eq:loss} using cosine similarity and $\tau=0.01$.
To balance low-resource languages, we sample each data batch from a multi-nominal distribution over different language subsets
\begin{equation}
\small
p_i = \frac{n_i^\alpha}{\sum_{j=1}^{n}n_j^\alpha},
\end{equation}
with $n_i$ equal to the size of subset $i$ and smoothing parameter $\alpha=0.5$.
We run the contrastive pre-training for 10k steps with a global batch size of 6192.
In-batch negatives are used and shared across different GPUs.
Each sequence is truncated at a maximum length of 128.
The learning rate is initially warmed up to 2e-4 for the first 10\% steps, followed by a linear decay.

We utilize the same contrastive loss during fine-tuning on each downstream dataset.
Each fine-tuning experiment only involves one dataset so we directly sample data after shuffling it.
Models are trained using a peak learning rate of 2e-5 with the same scheduler as pre-training.
The maximum sequence length is 512.
Batch size is 128 and each sample is accompanied with 7 randomly sampled negatives.
Training epochs is 3.
Other hyperparameters are same as pre-training.
We only consider in-batch negatives for contrastive learning so we compare models under this setting.

We conduct all experiments on two NVIDIA A100 GPUs with 40G memory.
We use DeepSpeed, gradient checkpointing and mixed precision (FP16) encoding to reduce memory cost.
The pre-training process takes about 18 hours. Fine-tuning on all datasets is finished in one day.

\subsection{Evaluation Benchmarks}

To provide an extensive evaluation of the generalization ability of our pre-trained models, we select a large variety of code retrieval tasks from different domains under different settings.

We first evaluate on the CodeSearchNet benchmark~\citep{husain2020codesearchnet}, which is widely used for evaluating the text-code search ability of code retrieval models.
One drawback of CodeSearchNet is the queries are not aligned to real user questions.
So, we also evaluate on some more challenging datasets, Adv Test~\citep{codexglue}, CoSQA~\citep{huang-etal-2021-cosqa}, CoNaLa~\citep{conala}, SO-DS~\citep{heyman2020neural}, StaQC~\citep{yao2018staqc}.
The last three evaluation datasets follow the setting of ~\citet{heyman2020neural}, where during inference both text description and code snippet are used for matching.
The main evaluation metric is MRR.

Then, code-code search results on POJ-104~\citep{poj14} is also reported to evaluate the intra-modal retrieval ability.
In this dataset, Python program solutions of the same problem is regarded as positive pairs.
The objective is to retrieve relevant code snippets which answer the same problem.
To investigate the cross-lingual code retrieval ability, we use CodeNet~\citep{puri2021codenet} as an evaluation benchmark, which is also a problem-solution dataset similar to POJ-104 but covers more languages.
On CodeNet, we consider the zero-shot retrieval of three programming languages (Ruby, Python and Java) following~\citet{guo-etal-2022-unixcoder}, where code is pre-processed by removing comments and replacing all separators with whitespace.
Performance is evaluated by MAP.

Finally, to test whether our model can generalize to out-of-domain languages, we choose two text-code search datasets with languages unseen during pre-training. Smart Contracts (SC)~\citep{solidity} contains Solidity programming language and Spider~\citep{yu-etal-2018-spider} consists of SQL-query pairs.
We use the dataset split released by ~\citet{chai2022cdcs}.
Models are evaluated by Recall@\{1,5,10\} and MRR@1000. The statistics of downstream evaluation benchmarks are illustrated in Table~\ref{tab:eval}.

\begin{table}[!ht]
\begin{tabular}{llrrr}
\toprule
Dataset    & Lang & Train & Valid & Test  \\
\midrule
CSN & Ruby & 24.9K    & 1.4K  & 1.3K  \\
CSN & JS & 58K    & 3.9K  & 3.3K  \\
CSN & Go & 167K   & 7.3K  & 8.1K  \\
CSN & Python & 252K   & 13.9K & 14.9K \\
CSN & Java & 165K   & 5.2K  & 10.9K \\
CSN & PHP & 241K   & 13.0K & 14.0K \\
\hline
Adv & Python & 28.0K   & 9.6K  & 19.2K \\
CoSQA & Python     & 19K    & 0.5K   & 0.5K   \\
CoNaLa & Python    & 2.8K     & -     & 0.8K   \\
SO-DS & Python    & 14.2K    & 0.9K   & 1.1K  \\
StaQC & Python    & 20.4K   & 2.6K  & 2.7K  \\
\hline
POJ104 & Python & 32K & 8K & 12K \\
\hline
CodeNet & Ruby & - & - & 11.7K \\
CodeNet & Python & - & - & 15.6K \\
CodeNet & Java & - & - & 23.5K \\
\hline
SC & Solidity & 57K & 4.1K & 1K \\
Spider & SQL & 14K & 2.1K & 1K \\
\bottomrule
\end{tabular}
\caption{Statistics of downstream evaluation datasets.}
\label{tab:eval}
\end{table}

\subsection{Results}

\begin{table*}[!ht]
\centering
\begin{tabular}{lccccccc}
\toprule
\textbf{Method}    & Ruby & Javascript & Go     & Python & Java  & PHP & Overall \\
\midrule
ContraCode~\cite{jain-etal-2021-contrastive}       & -                   & 30.6                & -               & -               & -               & -              & - \\
SyncoBERT~\cite{wang2021syncobert}        & 72.2                & 67.7                & 91.3            & 72.4            & 72.3            & 67.8           &74.0 \\
CodeBERT~\cite{feng-etal-2020-codebert}        & 67.9                & 62.0                & 88.2            & 67.2            & 67.6            & 62.8           &69.3 \\
GraphCodeBERT~\cite{guo2021graphcodebert} \hspace{-5pt}   & 70.3                & 64.4                & 89.7            & 69.2            & 69.1            & 64.9           &71.3 \\
UniXcoder~\cite{guo-etal-2022-unixcoder} &74.0&68.4&91.5&72.0&72.6&67.6&74.4 \\

CodeRetriever~\cite{li-etal-2022-coderetriever} &  75.3                &  69.5                &  91.6             &  73.3            &  74.0            &  68.2          &  75.3   \\
\hline
\ours & \textbf{77.8} & \textbf{72.5} & \textbf{92.4} & \textbf{76.1} & \textbf{75.7} & \textbf{70.1} & \textbf{77.4} \\
\bottomrule
\end{tabular}
\caption{MRR@1k on six programming language test sets of the CodeSearchNet.}
\label{tab:csn}
\end{table*}

\begin{table*}[!ht]
\centering
\begin{tabular}{lcccccc}
\toprule
            
\textbf{Method}  & Adv  & CoSQA & CoNaLa & SO-DS  & StaQC & Overall \\
\midrule
SyncoBERT~\cite{wang2021syncobert} & 38.1                & -               & -               & -               & -         & -      \\
CodeBERT~\cite{feng-etal-2020-codebert} & 27.2                & 64.7            & 20.9            & 23.1            & 23.4            & 31.9\\
GraphCodeBERT~\cite{guo2021graphcodebert} & 35.2                & 67.5            & 23.5            & 25.3            & 23.8           & 35.1 \\
UniXcoder~\cite{guo-etal-2022-unixcoder} & 41.3 & 70.1 &- & -& -& -\\
CodeRetriever~\cite{li-etal-2022-coderetriever} &  \textbf{43.0}  &  70.6  &  29.6       &  27.1 &  \textbf{25.5}  &  39.0 \\
\hline
\ours & 39.7 & \textbf{72.0} & \textbf{61.0} & \textbf{48.8} & 23.3 & \textbf{49.0} \\
\bottomrule

\end{tabular}
\caption{Text-code search performance (MRR@1k) on datasets that are closer to the real scenario.}
\label{tab:cs}
\end{table*}

In this section, we report and discuss the performance of \ours on the evaluation benchmarks introduced in the previous section, spanning both supervised and zero-shot settings.
In the supervised setting, \ours is directly fine-tuned on full training set and the last checkpoint is evaluated on the test set.
In the zero-shot setting, it is directly evaluated on the test set.

\paragraph{Text-Code Search}

We first present evaluation results on six programming language subsets of CodeSearchNet in Table~\ref{tab:csn}.
\ours trained with the newly proposed dataset outperforms previous best model CodeRetriever on all language subsets, by an average of 2.1 points.

Next, we look at results on several challenging benchmarks, all collected from real-world user queries instead of docstrings.
As shown in Table~\ref{tab:cs}, our model significantly outperforms prior state-of-the-art models by up to 10 points on average.
We attribute the improvement to real-world user queries from ProCQA.

\paragraph{Code-Code Search}

After evaluating the cross-modal search ability of our embedding model, we zoom into the intra-modal retrieval performance by evaluating on a code clone detection benchmark, POJ-104.
Results are illustrated in Table~\ref{tab:poj14}.
Our model outperforms previous best baseline CodeRetriever by +1.38 points.

\begin{table}[!ht]
\centering
\begin{tabular}{lc}
\toprule
\textbf{Method}         & MAP \\
\midrule
RoBERTa~\cite{roberta} & 76.67 \\
CodeBERT~\cite{feng-etal-2020-codebert} & 82.67 \\
GraphCodeBERT~\cite{guo2021graphcodebert} & 85.16 \\
SynCoBERT~\cite{wang2021syncobert} & 88.24 \\
DISCO~\cite{ding-etal-2022-towards} & 82.77 \\
Corder~\cite{corder} & 84.10 \\
CodeRetriever~\cite{li-etal-2022-coderetriever} & 88.85 \\
\hline
\ours & \textbf{90.23} \\
\bottomrule
\end{tabular}
\caption{Performance of Python code-to-code retrieval task on POJ-104.}
\label{tab:poj14}
\end{table}

\paragraph{Zero-Shot Cross-Lingual Code Search}

We list the cross-lingual code retrieval performance of our model \ours and other baselines from ~\citet{guo-etal-2022-unixcoder} in Table~\ref{table:cross_lingual_code_search}.
UniXCoder has significantly better zero-shot code retrieval performance, owing to its contrastive objective during pre-training.
\ours consistently outperforms previous baselines by a large margin, setting new state-of-the-art performance on this task.

\begin{table*}[!ht]
\centering
\resizebox{\textwidth}{!}{%
\begin{tabular}{lcccccccccc}
    \toprule
    \multirow{2}{*}{\textbf{Method}} & \multicolumn{3}{c}{Ruby} & \multicolumn{3}{c}{Python}& \multicolumn{3}{c}{Java} & \multirow{2}{*}{Overall} \\
    \cmidrule(l){2-4} \cmidrule(l){5-7} \cmidrule(l){8-10}
     & Ruby & Python & Java & Ruby & Python & Java & Ruby & Python & Java &\\
    \hline
    CodeBERT & 13.55 & 3.18 & 0.71 & 3.12 & 14.39 & 0.96 & 0.55& 0.42& 7.62& 4.94 \\
    GraphCodeBERT \hspace{-5pt} & 17.01 & 9.29 & 6.38 & 5.01 & 19.34 & 6.92 & 1.77& 3.50& 13.31& 9.17\\
    PLBART & 18.60 & 10.76 & 1.90 & 8.27 & 19.55 & 1.98 & 1.47& 1.27& 10.41& 8.25 \\
    CodeT5 & 18.22 & 10.02 & 1.81 & 8.74 & 17.83 & 1.58 & 1.13& 0.81& 10.18& 7.81 \\
    UniXcoder & 29.05 & 26.36 & 15.16 & 23.96 & 30.15 & 15.07 & 13.61 & 14.53 & 16.12 & 20.45 \\
    \hline
    \ours & \textbf{44.74} & \textbf{43.11} & \textbf{31.26} & \textbf{40.59} & \textbf{45.77} & \textbf{29.75} & \textbf{32.8} & \textbf{33.75} & \textbf{30.74} & \textbf{36.95} \\
    \bottomrule
\end{tabular}
}
\caption{MAP score (\%) of zero-shot code-to-code search task on CodeNet.}
\label{table:cross_lingual_code_search}
\end{table*}

\paragraph{Cross-Domain Code Search}
In Table~\ref{tab:cdcs}, we compare different models' performance on Solidity and SQL, two languages unseen during pre-training.
Previous best model MAML~\citep{chai2022cdcs} applied model-agnostic meta learning on CodeBERT~\citep{feng-etal-2020-codebert} using Java and Python subsets from CSN for pre-training.
In addition, we also report the performance of GraphCodeBERT using our codebase as another baseline for comparison.
Our model significantly improves the cross-domain code search performance on unseen languages.
One possible reason is that the diversity of language coverage in ProCQA equips the model with better language adaptation ability.

\begin{table*}[!ht]
\centering
\begin{tabular}{lcccccccc}
\toprule
\multicolumn{1}{c}{\multirow{2}{*}{\textbf{Method}}} & \multicolumn{4}{c}{Solidity} & \multicolumn{4}{c}{SQL} \\ \cmidrule(l){2-5} \cmidrule(l){6-9} 
\multicolumn{1}{c}{} & R@1 & R@5 & R@10 & MRR & R@1 & R@5 & R@10 & MRR \\ \midrule
CodeBERT~\cite{feng-etal-2020-codebert} & 53.2 & 77.9 & 84.8 & 64.4 & 67.5 & 92.0 & 96.0 & 78.2 \\
MAML~\cite{chai2022cdcs} & 65.8 & 82.9 & 87.9 & 73.4 & 74.6 & 95.2 & 97.2 & 83.7 \\ 
GraphCodeBERT~\cite{guo2021graphcodebert} \hspace{-5pt} & 72.9 & 85.5 & 89.3 & 78.5 & 78.5 & 94.5 & 96.6 & 85.5 \\ \hline
\ours & \textbf{75.2} & \textbf{87.3} & \textbf{90.6} & \textbf{80.7} & \textbf{85.4} & \textbf{96.0} & \textbf{97.4} & \textbf{90.3} \\
\bottomrule
\end{tabular}
\caption{
Results of cross-domain code retrieval on programming languages unseen during pretraining.}
\label{tab:cdcs}
\end{table*}

\subsection{Analysis}

\begin{figure}[!ht]
    \centering
    \includegraphics[width=0.48\textwidth]{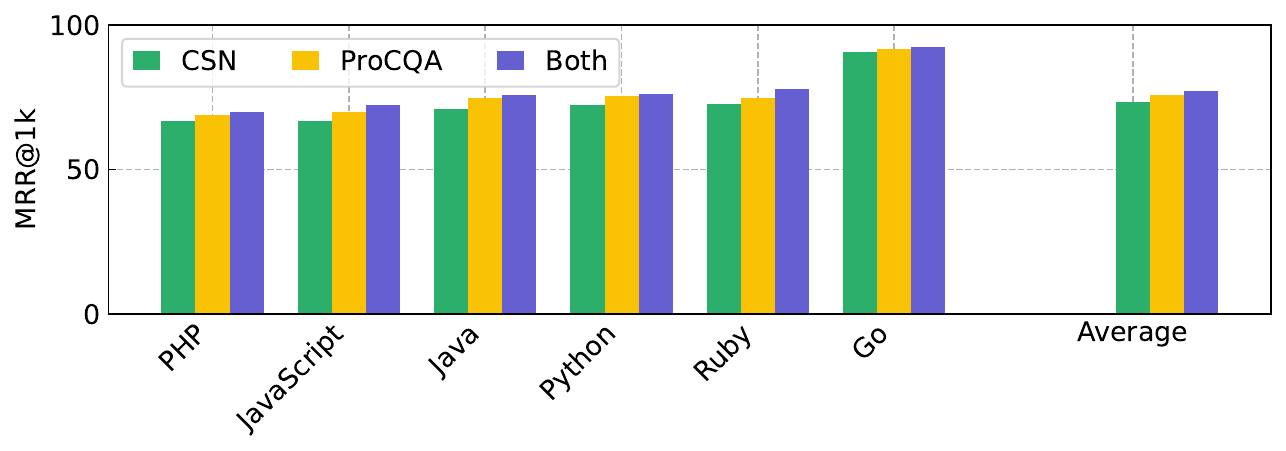}
    \caption{Ablation of the pre-training corpus. Results compared on test sets of CodeSearchNet.}
    \label{fig:data_ablation}
\end{figure}

\paragraph{Impact of pretraining data distribution}
We first investigate the effect of different pre-training corpus by doing a series of controlled experiments where only the pre-training data distribution is changed.
We run two additional experiments by using the CSN and ProCQA dataset individually for pre-training.
Due to space limitation, we report downstream fine-tuned retrieval performance on CodeSearchNet in Figure~\ref{fig:data_ablation}.
Results on other evaluation benchmarks follow the same trends.

Despite CSN belongs to in-domain data for this evaluation benchmark, it still underperforms ProCQA when being used as a pre-training corpus.
Combining two datasets gives better results.
This showcases the effectiveness of ProCQA dataset being used as a mixed-modal corpus for retrieval-oriented contrastive pre-training.

\paragraph{Effect of modality-agnostic data}
We ablate on the choice of modality-agnostic contrastive learning by comparing to another setting which we explicitly distinguish text and code in data design.
Due to the high difficulty of parsing incomplete code snippets in ProCQA, we conduct this ablation on the CSN pre-training corpus where code are well formed and can be parsed by existing tools. Bi-modal setting removes all comments in the code while mixed-modal setting keeps them.
We list results in Table~\ref{tab:dataformat}.
The evaluation set Adv Test only requires code-text matching, yet training with mixed-modal data formats still has benefits.

\begin{table}[!ht]
    \centering
    \begin{tabular}{rc}
    \toprule
    Setting & Adv Test \\
    \midrule
    bi-modal     & 38.8 \\
    mixed-modal & 39.1 \\
    \bottomrule
    \end{tabular}
    \caption{Effect of the data format used in CodeSearchNet corpus pre-training. We report MRR@1k on Adv Test set.}
    \label{tab:dataformat}
\end{table}

\paragraph{Quantifying the effect of data contamination}

To avoid data contamination and ensure fairness, we performed de-duplication for the ProCQA dataset with respect to the relevant evaluation benchmarks from the same source, including CoNaLa, SO-DS and StaQC.
In Table~\ref{tab:decontamination}, we provide a quantitative analysis on the proportion of contaminated data for each evaluation set and the performance using raw and filtered version of the ProCQA dataset for pre-training.
Although a large-proportion of the evaluation set is included in the raw pre-training data, removing them raises a limited degradation of model performance, as they only make up a small portion of the large-scale pre-training data.

\begin{table}[!ht]
    \centering
    \begin{tabular}{lccc}
    \toprule
    Dataset & Proportion & Unfiltered & Filtered \\
    \midrule
    CoNaLa & 83.7\% & 62.9 & 61.0 \\
    SO-DS & 48.4\% & 49.8 & 48.8 \\
    StaQC & 31.4\% & 23.6 & 23.3 \\
    \bottomrule
    \end{tabular}
    \caption{Analysis on the influence of data contamination on three evaluation datasets.}
    \label{tab:decontamination}
\end{table}

\section{Conclusion}
In this work we introduce ProCQA, a large-scale community-based programming question answering dataset mined from StackOverflow with strict filtering strategies for quality and fairness control.
ProCQA is featured by its practicality, diversity and code-mixing data format.
Furthermore, through modality-agnostic contrastive pre-training on interleaved code and text data, our new dataset yields a language model that has a better aligned representation space between code and text, achieving state-of-the-art performance on a large spectrum of code retrieval tasks.
In future work, it would be interesting to explore the benefit of ProCQA to other generative code QA tasks.

\section{Acknowledgements}
This work was supported by the National Natural Science Foundation of China (No.61977003) and the State Key Laboratory of Complex \& Critical Software Environment (CCSE-2024ZX-16).

\balance
\nocite{*}
\section{Bibliographical References}\label{sec:reference}
\bibliographystyle{lrec-coling2024-natbib}
\bibliography{lrec-coling2024-example}


\end{document}